\documentclass[lettersize,journal]{IEEEtran}
\usepackage{amsmath,amsfonts}
\usepackage{algorithmic}
\usepackage{algorithm}
\usepackage{array}
\usepackage[caption=false,font=footnotesize,labelfont=rm,textfont=rm]{subfig}
\usepackage{textcomp}
\usepackage{stfloats}
\usepackage{url}
\usepackage{verbatim}
\usepackage{graphicx}
\usepackage{cite}

\usepackage{multirow}
\usepackage{booktabs}
\usepackage{makecell}
\usepackage{amsmath}
\usepackage{soul,color,xcolor}
\usepackage{cleveref}
\begin{document}

\title{Dual-Attention Heterogeneous Graph Neural Network for Multi-robot Collaborative Area Search via Deep Reinforcement Learning}

\author{Lina Zhu, Jiyu Cheng, Yuehu Liu, and Wei Zhang,~\IEEEmembership{Senior Member,~IEEE}
\thanks{Lina Zhu is with the School of Artificial Intelligence, School of Automation, Xi'an University of Posts \& Telecommunications, Xi'an, China, 710121 (e-mail: linazhu@xupt.edu.cn).

        Yuehu Liu is with the Institute of Artificial Intelligence and Robotics, Xi’an Jiaotong University, Xi’an, China, 710049 (e-mail: liuyh@xjtu.edu.cn).
        
        Jiyu Cheng and Wei Zhang are with the School of Control Science and Engineering, Shandong University, Shandong, China, 250061 (e-mail:  { \{jycheng, davidzhang\}@sdu.edu.cn}).
        
        \textit{(Corresponding author: Jiyu Cheng.)}}}

\markboth{}%
{Zhu \MakeLowercase{\textit{et al.}}: Dual-Attention Heterogeneous GNN for Multi-robot Collaborative Area Search via Deep Reinforcement Learning}


\maketitle

\begin{abstract}
In multi-robot collaborative area search, a key challenge is to dynamically balance the two objectives of exploring unknown areas and covering specific targets to be rescued. Existing methods are often constrained by homogeneous graph representations, thus failing to model and balance these distinct tasks. To address this problem, we propose a Dual-Attention Heterogeneous Graph Neural Network (DA-HGNN) trained using deep reinforcement learning. Our method constructs a heterogeneous graph that incorporates three entity types: robot nodes, frontier nodes, and interesting nodes, as well as their historical states. The dual-attention mechanism comprises the relational-aware attention and type-aware attention operations. The relational-aware attention captures the complex spatio-temporal relationships among robots and candidate goals. Building on this relational-aware heterogeneous graph, the type-aware attention separately computes the relevance between robots and each goal type (frontiers vs. points of interest), thereby decoupling the exploration and coverage from the unified tasks. Extensive experiments conducted in interactive 3D scenarios within the iGibson simulator, leveraging the Gibson and MatterPort3D datasets, validate the superior scalability and generalization capability of the proposed approach.

\end{abstract}


\section{Introduction}
    \IEEEPARstart{M}{ulti-robot} area search (MAS) is a fundamental research problem in multi-robot systems, focused on how multiple robots can efficiently rescue trapped people or targets in unknown environments. MAS has broad applications, including Mars exploration \cite{nilsson2018toward}, disaster response \cite{liu2016multirobot}, and urban search and rescue \cite{hong2019investigating,shree2021exploiting}. To achieve effective rescue, robots' decision-making must encompass several key capabilities. Firstly, to explore the unknown environment to gather information about potential targets (exploration). Secondly, to cover (or localize) the targets within the explored area (coverage). Finally, there is a trade-off between exploration and coverage for maximizing the number of rescued targets, which means it should consider discovering more targets in unexplored areas or rescuing more targets in explored areas.
    
    Compared to the occupancy grid map, the topological maps offer significant advantages in capturing the spatial relationships between the robots and sub-task relevant locations, specifically concerning frontier (for exploration) or the points of interest (for coverage) in multi-robot area search tasks\cite{zhang2022h2gnn}. Moreover, the topological map is less sensitive to environmental changes and does not depend on fixed spatial resolutions, making it highly effective and scalable in dynamic and large-scale environments. Therefore, the end-to-end deep reinforcement learning (DRL) methods that directly map from the embedding vector of the topological map to the actions catalyze the development of sequential decision-making tasks\cite{c55}\cite{c56}. However, as the number of nodes on the graph changes or increases, the state and action spaces expand, leading to potential instability in training and suboptimal solutions in policies trained through end-to-end DRL methods. We employ a hierarchical policy that divides the decision-making process into global and local planning phases. The upper-level global planning of upper-level policy determines the long-term goal by selecting the candidate nodes that maximize the cumulative reward. The local planner then guides the robots toward the assigned long-term goal.
    
    The hierarchical policy mentioned above enables simultaneous execution of the exploration and coverage sub-tasks, which allows for better coordination between the two sub-tasks \cite{tolstaya2021multi}. This unified approach also significantly reduces the computational cost compared to handling two sub-tasks independently. Specifically, the robot nodes on the topological cross-graph simultaneously select both the frontier nodes and interesting nodes. Considering the weight of two sub-tasks during the selection process is well-motivated and will further enhance efficiency. For instance, in the initial phase, the robots are inclined to explore larger areas, as the points of interest are often hidden in the unexplored areas. As the explored areas expand and more points of interest are exposed, the robots should focus on covering these newly discovered points of interest. In \cite{zhu2023autonomous}, the trade-off between exploration and coverage is adjusted by the various reward weight coefficients, which rest in the number distribution of points of interest. In \cite{patil2023graph}, Patil \emph{et al.} take the distance cost, information utility, information gain, and the self-defined importance weight into account to allocate the goals to the robots. These manually designed heuristic methods exhibit poor adaptive ability and scalability, limiting their effectiveness in dynamic or larger-scale scenes. In our work, we autonomously learn to balance exploration and coverage, dynamically adapt task priority based on the environment state and task requirement.

\begin{figure*}[!t]
    \centering
    \includegraphics[scale=0.45]{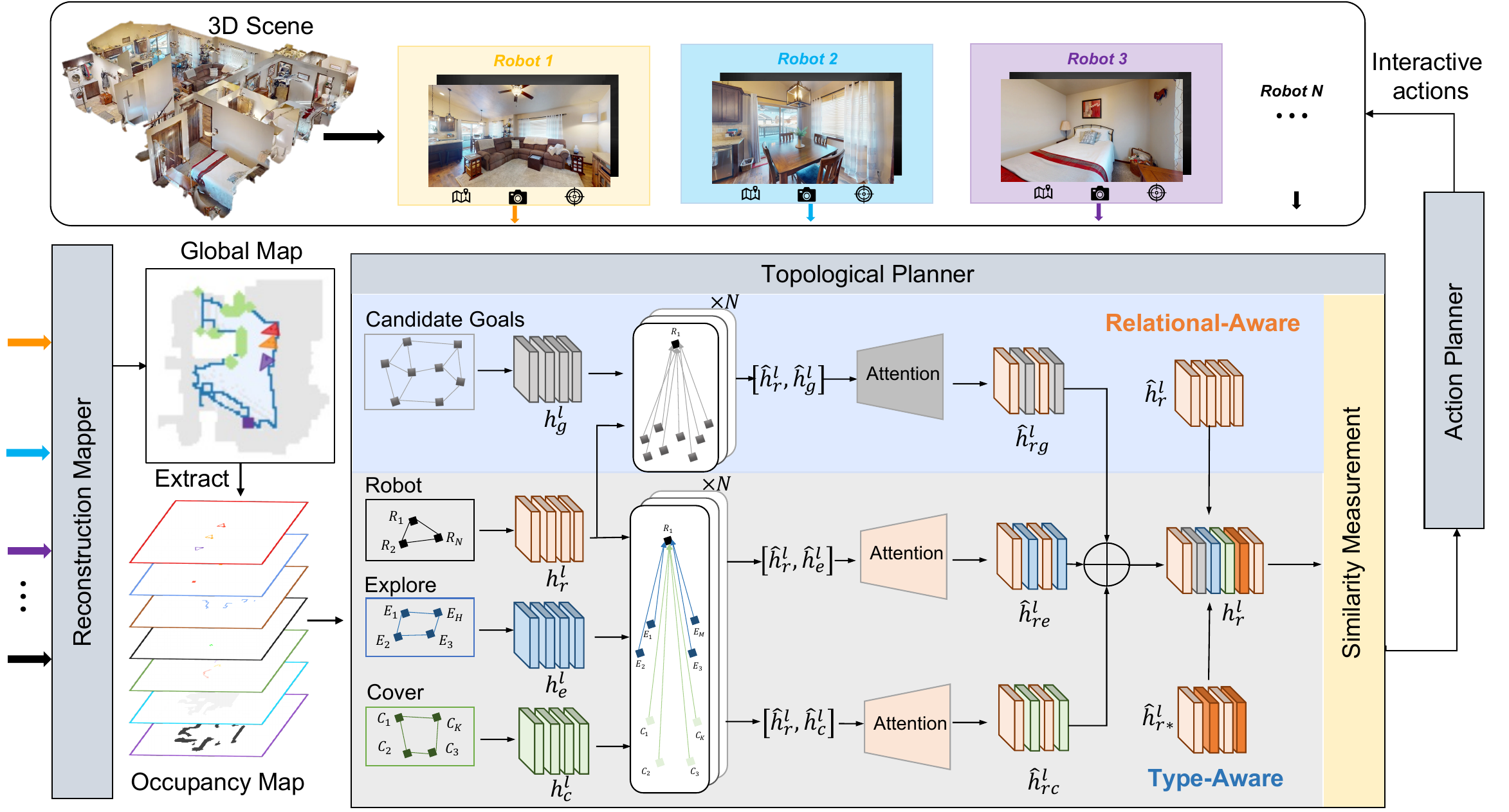}
    \caption{The overall framework of our method. The multi-robot area search method comprises three modules: a Reconstruction mapper, a Topological planner, and an Action planner. The topological planner is responsible for assigning a long-term goal to each robot over a heterogeneous graph. This is achieved through a structured three-stage process: Relational-aware graph construction, Type-aware graph augmentation, and Similarity measurement. }
    \label{fig:framework}
    \end{figure*}

    In summary, we propose a Dual-Attention Heterogeneous Graph Neural Network (DA-HGNN), a unified decision-making framework that operates on a constructed heterogeneous graph, which enables both sub-tasks exploration and coverage to be executed simultaneously. The dual-attention mechanism dynamically evaluates and integrates node importance from two dimensions: relational‑aware attention and type‑aware attention. Specifically, the model first constructs a relation‑aware spatio‑temporal graph about the robots and the other entities, then augments it with explicit type‑aware representations about the two types of sub-tasks. Finally, goal assignment is formulated as a similarity‑measurement problem, where the optimal matching between robots and candidate goals is learned by maximizing node correspondences over the enriched graph. All illustrated in Fig. \ref{fig:framework}, this approach enables robots to hierarchically integrate scene information and autonomously adjust policy based on the state, achieving adaptive and coordinated multi‑robot search.
    
    Our main contributions are as follows:

    (1) We propose a Dual-Attention Heterogeneous Graph Neural Network (DA-HGNN) that explicitly models robot–goal interactions through relational-aware and type-aware attention mechanisms, enabling dynamic balanced decision-making between exploration and coverage tasks.
    
    (2) We employ a hierarchical architecture for multi-robot area search tasks, where the goal assignment problem is formulated as a similarity measurement across multiple graphs during the upper-level planning phase.
    
    (3) We demonstrate the scalability and generalization of our proposed method compared to the state-of-the-art methods through well-designed experiments in visually and physically interactive 3D scenes, conducted in the iGibson simulator with Gibson and MatterPort3D datasets. 


\section{Related Work}

    The multi-robot coordinated area search problem has attracted considerable research interest due to its broad range of applications. Existing studies primarily address it from three aspects: optimization-based methods, traditional heuristic-based methods, and learning-based methods.

    The optimization-based method formulates the area search problem as a multi-robot routing and scheduling problem, aiming to minimize the time and energy consumption of the robots. The solution provides the optimal routes or an efficient schedule for all robots \cite{lin2025high,es1689}. For example, Chen \emph{et al.} \cite{10964773} proposed a mathematical model that incorporates realistic constraints for the multi-objective multi-robot routing and scheduling problem. Romeh \emph{et al.} \cite{el2025multi} integrated the deterministic task allocation of multi-robot exploration with the multi-objective slap swarm algorithm to balance exploration efficiency and mapping accuracy. Kim \emph{et al.} \cite{kim2025coordinated} utilized the traveling salesman problem to calculate the minimum exploration paths in the located area of each robot. While optimization-based methods can guarantee optimality in small scenes with a limited robot, they usually face significant challenges in computational complexity.

    Traditional heuristic-based methods usually allocate a sub-area to each robot and then optimize the coverage for each sub-area \cite{11149758}. Cao \emph{et al.} \cite{cao2025multi} introduced a dynamic coverage strategy where robot assignments are continuously optimized online by the centroid Voronoi tessellation. Q. Dong \emph{et al.} \cite{dong2024fast} presents a multi-UAV exploration strategy that utilizes a dynamic topological graph and graph Voronoi partition to achieve task allocation. Das, Sanath Kumar \emph{et al.} \cite{11022590} employed a dynamic PID-controlled area partitioning strategy and TSP-based path planning within unknown convex polygonal scenes. Although the heuristic methods of multi-robot search achieve uniform partitioning through effective resource allocation, their robustness depends on the model of scenes, and they fail to generalize to novel or unseen environments.

    The learning-based methods directly learn planning actions from environmental observations by the decision-making policy, which mitigates computational complexity while enhancing scalability and generalization compared with optimization-based and heuristic-based methods \cite{zhang2022h2gnn,zhang2024nowhere}. Furthermore, the hierarchical frameworks have been introduced to reduce redundant exploration efficiency and improve the coverage efficiency \cite{zhu2024multi,zhu2025autonomous}. In such frameworks, the high-level module first obtains scene information through the exploration algorithm and calculates the long-term goal, while the low-level module plans the robot's path from its current position to the goal. Despite the learning-based methods achieving autonomous planning and simultaneously executing the exploration and coverage in a unified process, these methods overlook the relative priority of the two sub-tasks. In this work, we incorporate a sub-task importance evaluation into a hierarchical policy, achieving the adaptive trade-off between the two sub-tasks.

\section{Problem Formulation}

    In multi-robot area search tasks involving two key sub-tasks: exploration and coverage, each robot at every timestep $t$ uses onboard sensors and RGB-D cameras to capture egocentric local observations and pose information from the 3D scene. Based on the occupancy map, a global policy selects a long‑term goal for each robot from a set of candidate goals, which includes the frontier (for area exploration) and points of interest (for target coverage). This long‑term goal serves as the directional guidance for a global planning cycle. Each global cycle spans $k$ timesteps of local motion planning, during which the robot follows a collision‑free path from its current location toward its assigned long‑term goal. The planned path is further converted into low‑level motion. At each timestep within the local planning, the robot receives environment rewards for both exploration and coverage. The cumulative reward collected over the $k$‑step is used for training and updating the global policy to maximize the long‑term cumulative return for sustained policy improvement.
    

\section{Modeling of Markov Decision Process}

     The upper-level policy in a hierarchical architecture for the multi-robot area search task aims at the long-term goals assignment for each robot. Considering the scene with reliable communication, the centralized decision-making policy is adopted. In this setting, the multi-robot area search problem can be formulated as a decentralized Partially Observable Markov Decision Process (POMDP). POMDP is defined by the tuple: $<\mathcal{N},\mathcal{S}, \mathcal{A}, \mathcal{P}, \mathcal{R}, \mathcal{Z}, \mathcal{O}, \gamma>$. We consider a team of $\mathcal{N}$ robots, indexed by $\mathcal{N}=\{1,2,\dots, N\}$. The global state $s \in \mathcal{S}$ evolves as robots execute joint action $a = (a^1,\cdots, a^N) $ from the joint action space $\mathcal{A} := \mathcal{A}^1 \times \dots \times \mathcal{A}^N$. The state transitions $s^{\prime} \in \mathcal{S}$ are governed by probability transition function $\mathrm{P}\left({s}^{\prime}\mid{s}, {a}\right)$: $\mathcal{S} \times \mathcal{A} \to \mathcal{S}^{\prime}$. The team reward $\mathcal{R}$ is determined by function $r(s,{a})$: $\mathcal{S} \times \mathcal{A} \times \mathcal{S}^{\prime} \to \mathbb{R}$, and ${o}^i \in \mathcal{Z}^i$ denotes the local observation generated by a global state $s$ and a local observation function $\mathcal{O}(s,i)$ : $\mathcal{S} \times \mathcal{N} \to \mathcal{Z}$. Each robot $i$ bases its learns policy $\pi^i(a^i|\tau^i)$ : $ \mathcal{T} \times \mathcal{A} \to [0,1]$ on its local action-observation history $\tau^i \in \mathcal{T} \equiv (\mathcal{Z}^i \times \mathcal{A}^i)$. The objective of all robots is to jointly optimize policies to maximize the expected discounted cumulative return $\sum_{t=1}^{T}\gamma^{t}r_t$, where $\gamma \in [0,1]$ is the discount factor.
    
\section{Methodology} 
    In this section, we introduce the overall framework of our algorithm. The entire pipeline consists of three key modules: Reconstruction mapper, Topological planner, and Action planner, as shown in Fig. \ref{fig:framework}. The Reconstruction mapper module predicts the top-down global map based on the allocentric image observations and positions of multiple robots. The Topological planner takes the global occupancy map to construct a Dual-Attention Heterogeneous Graph to model robot-goal relationships. This is achieved through three sub-models: the Relational-aware Graph Construction builds the spatio-temporal graph among robots and goal candidates, the Type-aware Graph Construction augmentation enriches this graph by explicitly distinguishing between the frontier and interesting nodes, and then the Similarity Measurement sub-module evaluates candidate goals and assigns each robot a long-term goal by integrating both task-specific and relational contexts. Once the topological planner updates at each global timestep, the Action Planner computes a feasible path from the robot's current location to its assigned long-term goal and converts it into a sequence of interactive actions. The action planner subsequently guides the robots in exploring unknown areas and covering target over the following $k$ local timesteps.

\subsection{Reconstruction mapper}

    We introduce the reconstruction mapper to provide a top-down view global map from RGB-D images and global positions of all robots, as shown in Fig. \ref{fig:framework}. The calculation process for the top-down view global map is to back-project the depth image from the 2D plane as 3D point cloud data, then the point cloud data is compressed onto the 2D plane. Specifically, the robots obtain the allocentric image observation $o_t$ from the 3D space scene $S_t$. Then, provided camera intrinsic and global positions, each pixel value in the allocentric depth image is back-projected into the 3D plane to generate the cloud point data $P_t=\{p_k|k=1,2,\dots,h\times w\}$. The back-projection principle involves geometrically transforming the point from the image coordinate system to the world coordinate system using the camera intrinsic. From the 3D cloud point data to the global map, this compression process considers the height relation between the 3D point and the robots. If the value of 3D point $p_k$ is larger than 0 and less than the robot's height $h_r$, the cell corresponding to the coordinate of 3D point in the global map is defined as occupied, which means this cell is the obstacle. The cell in the global grid map can be classified into one of the four categories: occupied (obstacle), open (explored but not occupied), unlocated (explored but not occupied and located), and unknown (unexplored). The unlocated cell indicates that a target falls into this cell, and the robots do not cover (or locate) the target. The global grid map at timestep $t$ is represented as $M_t \in [0,1]^{X \times Y \times 3}$, where $X, Y$ are the map size. Three grid map channels indicate the explored, uncovered, and occupied regions.


\subsection{Topological Planner}
    The Topological Planner serves as the core decision-making module for multi-robot collaborative search, designed to assign a long-term goal to each robot by reasoning over the structural and semantic information of the scene. It consists of three integrated sub-modules: Relational-Aware Graph Construction, Type-aware Graph Construction, and Similarity Measurement. The relational-aware graph construction builds a spatio-temporal relational graph among robots and candidate goals. The type-aware graph augmentation enriches the task semantics within the relational graph by explicitly distinguishing between frontier nodes (for exploration) and interesting nodes (for coverage), thereby forming a heterogeneous graph that jointly encodes both relational and semantic information. Then, the Similarity Measurement evaluates the correspondence between robots and candidate goals on the augmented heterogeneous graph, ultimately assigning the most beneficial goal for each robot as the long-term goal. 

\subsubsection{Relational-Aware Graph Construction}

     The core of multi-robot collaboration decision-making is to capture the spatio-temporal relationships between robots and all entities of the environment. 
     This sub-module is designed to capture the essential spatio-temporal relationships for multi-robot collaborative decision-making. Relying solely on individual robots' local observation is insufficient for achieving effective coordination. To address this, we construct self-graphs for the robots, candidate goals, and their historical states, and further establish cross-graphs that encode topological relationships between different entities. This process provides the relational infrastructure and serves as the structural foundation for our heterogeneous graph.
     
     The top-down view global grid map $M_t$ is generated by the reconstruction map. Then extract the environment occupancy map layer-by-layer from $M_t$, which contains the robots, frontiers, targets, obstacles, and explored channels, denoted as $o_t^e \in [0,1]^{X \times Y \times 5} $. Then the observation map is expanded with the historical positions of all robots $o_t^{hp} \in [0,1]^{X \times Y} $, and historical goals $o_t^{hg} \in [0,1]^{X \times Y}$, enabling the observation map is gradually approximate the state, contributing to the environment representation, and avoiding the robot redundant moving the located frontier or targets noses. Therefore, the observation information is updated as the 7-channel map and denoted as $o_t \in [0,1]^{X \times Y \times 7}$, which indicates the robots, historical position of all robots, frontiers, targets, obstacles, explored region, and historical goals. We design the convolutional neural network to extract the high-dimensional feature vector from the expanded observation map. Specifically, first, the occupancy map is downsampled to reduce its spatial resolution and data volume. The observation information with the size of $X \times Y \times 7$ is downsampled to decrease the spatial dimension and result in a lower-resolution map  $o_t^{\prime}$, reducing the information size to $X_s \times Y_s \times 7$, the $X_s$ and $Y_s$ is set to $X/4$ and $Y/4$ in our method, respectively. Then we encode the lower-resolution observation map $o_t^{\prime}$ as the high-dimension feature vector $h \in \mathbb{R}^{7 \times 64}$ via the multi-layer perception network. Each channel map of 7 channels in the occupancy map is embedded into a 64-dimensional feature vector and preserves the critical information.

     After obtaining the feature vector of each entity, which is set as the node to construct the graph, the relationship between itself and its neighbors' nodes is aggregated. The overall graph neural network is composed of four self-graphs and three cross-graphs. The four self-graphs including the robot-info graph $G_r = \{V_r, E_r\}$, candidate-goal-info graph $G_g = \{V_g, E_g\}$, historical-robot-info graph $G_{r^{\prime}} = \{V_{r^{\prime}},E_{r^{\prime}}\}$, historical-goal-info graph $G_{g^{\prime}} = \{V_{g^{\prime}},E_{g^{\prime}}\}$. The node $V_r$ of the robot-info graph is set as the feature vector of all robots $h_r \in \mathbb{R}^{64}$, and the edge feature $E_r$ is the geometric distance between the robot and the others. Meanwhile, the setting of node and edge features of candidate-goal-info, historical-robot-info, and historical-goal-info graph are similar to the robot-info graph. 

     Each node of the self-graph performs an aggregation operation on the node of local neighborhoods via the multi-head attention mechanism \cite{vaswani2017attention}, which has been shown can stabilize the policy learning process and enhance the model capacity \cite{velivckovic2018graph}. We have the cold start feature $h_{r\_i}^0 \in \mathbb{R}^{64}$ for per-node $i \in V_r$ of the robot-info graph. An attention head $m$ in the $l$-th layer of node feature updates the corresponding hidden feature $h_{r\_i}^{(l,m)}$ with its neighborhood. We first apply a trainable weigh parameter $W_{r\_ij}^{(l,m)}$ to each neighboring node $j \in \mathcal{N}_i$ of node $i$:

     \begin{equation}
        W_{r\_ij}^{(l,m)} = \frac{exp(k_{r\_i}^{(l,m)} \cdot q_{r\_j}^{(l,m)}) }{\sum\limits_{u \in N_i} exp(k_{r\_u}^{(l,m)} \cdot q_{r\_i}^{(l,m)})}
     \label{equ_6}
     \end{equation}
     where $k_{r\_i}^{(l,m)}$, $q_{r\_j}^{(l,m)}$ are the key and query for attention mechanism, which are computed by the $k_{r\_i}^{(l,m)} = W_k^{(l,m)} \cdot h_{r\_i}^{(l,m)}$, and $q_{r\_i}^{(l,m)} = W_q^{(l,m)} \cdot h_{r\_i}^{(l,m)}$. This calculating process is the normalization of all weight coefficients.

     The node feature $h_{r\_i}^{(l)}$ at layer $l$ is calculated by concatenating the weigh of all head $M$,    
     
     \begin{equation}
         h_{r\_i}^{(l)} = \underset{m=1}{\overset{M}{\|}} \, h_{r\_i}^{(l, m)} 
     \end{equation}     
     
     Then the node feature $h_{r\_i}^{(l+1)}$ at layer $l+1$ is aggregated the feature vector of node $i$ and neighbors' feature vector,

     \begin{equation}
        h_{r\_i}^{(l+1)} = h_{r\_i}^{(l)} + \rho_v([h_{r\_i}^{(l)} \| \sum\limits_{u \in \mathcal{N}_i} W_{r\_iu}^{(l)} \cdot h_{r\_u}^{(l)}]).
     \label{equ_7}
     \end{equation}
     where $\rho_v(\cdot)$ is the feature aggregation function based on the multi-layer perception, $[\cdot \| \cdot]$ denotes the concatenation.

     Meanwhile, the aggregation operation of the edge feature, the node feature of $G_g$, $G_r^{\prime}$, and $G_g^{\prime}$ are similar to the above-mentioned updated process of the node feature.
     
     The aggregated node feature of the self-graph is used for the initial state of the node feature of the cross-graph $G_{rg} = \{V_{rg}, E_{rg}\}$ between the robot and candidate goals. The distance between the robot and the candidate goal is calculated by the Fast Marching Method, which is considered in the aggregation of edge features. Then the attention weigh parameter $W_{rg\_ij}^{(l,m)}$ of node $i$ over neighbors node $j$ as below,

     \begin{equation}
        W_{rg\_ij}^{(l,m)} = \frac{exp(\rho_v(k_{rg\_i}^{(l,m)} \| q_{rg\_j}^{(l,m)} \| d_{rg_{i},rg_{j}})) }{\sum\limits_{u \in N_i} exp(\rho_v(k_{rg\_u}^{(l,m)} \| q_{rg\_i}^{(l,m)} \| d_{rg_{i},rg_{u}}))}
     \label{equ_6}
     \end{equation}
     where $rg_{i}$ denotes the node $i$ of cross-graph $G_{rg}$, $ d_{rg_{i},rg_{j}}$ denotes the distance between node $rg_{i}$ and $rg_{j}$.

     Then the node feature $h_{rg\_i}^{(l+1)}$ at layer $l+1$ is aggregated the feature vector of node $rg_{i}$ and neighbors' feature vector,

     \begin{equation}
        h_{rg\_i}^{(l+1)} = h_{rg\_i}^{(l)} + \rho_v([h_{rg\_i}^{(l)} \| \sum\limits_{u \in \mathcal{N}_i} W_{rg\_iu}^{(l)} \cdot h_{rg\_u}^{(l)}]).
     \label{equ_10}
     \end{equation}

\subsubsection{Type-aware Graph Augmentation}
    
    Building upon the relational-aware graph that captures spatio-temporal relationships through topological-level attention. The primary representation of spatial relationships between robots and candidate goals, without distinguishing between frontier nodes (for exploration) and interesting nodes (for coverage). This limits decision-making ability to dynamically weigh which goal type is more beneficial for long-term planning. To explicitly enhance task semantics, we introduce a type-aware graph augmentation sub-module. Specifically, we construct two more self-graphs over the explore and target nodes, denoted as the $G_e = \{V_e, E_e\}$ and $G_c = \{V_c, E_c\}$, respectively. We construct four cross-graphs, including robot-explore-info graph $G_{re}$, robot-cover-info graph $G_{rc}$, historical-robot-explore-info graph $G_{r{\prime}e}$, historical-robot-cover-info graph $G_{r{\prime}c}$. Firstly, the feature vector $h_{e\_i}^l$ of node $i$ in the self-graph $G_e$ at layer $l$ is aggregated with neighboring feature vector:

     \begin{equation}
        h_{e\_i}^{(l)} = \rho_v([h_{e\_i}^{(l)} \| \sum\limits_{u \in \mathcal{N}_i} W_{e\_iu}^{(l)} \cdot h_{e\_u}^{(l)}]).
     \label{eqa:1}
     \end{equation}
     where $W_{e\_iu}^{(l)}$ denotes the self attention parameters between node $i$ and neighboring node $u$, the $h_{e\_u}^{(l)}$ denotes the feature vector of node $u$. 

     Then the feature vector of node $re\_i$ on cross-graph $G_{r\_e}$ is propagated to node $re\_i$ multiple times with corresponding attention weight $W_{re\_iu}^{(l)}$. The attention operation process of the robot and frontier as shown in Fig. \ref{fig:framework}, is denoted as $[\hat{h}_r^l,\hat{h}_e^l]$.

     \begin{equation}
     \begin{array}{l}
         \hat{h}_{re\_i}^{(l)} = \rho_v([\hat{h}_{re\_i}^{(l)} \| \sum\limits_{u \in \mathcal{N}_i} W_{re\_iu}^{(l)} \cdot \hat{h}_{re\_u}^{(l)}]) \\
        \vspace{0.3cm}        
         \hat{h}_{rc\_i}^{(l)} = \rho_v([\hat{h}_{rc\_i}^{(l)} \| \sum\limits_{u \in \mathcal{N}_i} W_{rc\_iu}^{(l)} \cdot \hat{h}_{rc\_u}^{(l)}])
    \end{array}
     \label{eqa:2}
     \end{equation}
    where the $\hat{h}_*^{(l)}$ denotes the aggregated feature vector by multi-head attention operation.

    The feature vector of robot $r$ on self-graph $\hat{h}_{r}^{(l)}$ is enhance by adding the type-aware feature $\hat{h}_{re}^{(l)}$ and $\hat{h}_{rc}^{(l)}$.

    \begin{equation}
        {h}_{r}^{(l)} = \hat{h}_{r}^{(l)} +  \hat{h}_{rg}^{(l)} + \hat{h}_{re}^{(l)} + \hat{h}_{rc}^{(l)} + h_{rr^{\prime}}^{(l)} + \hat{h}_{r*}^{(l)} .
    \end{equation}

     \begin{figure}[!t]
        \vspace{0.5em}
        \centering
        \includegraphics[scale=0.6]{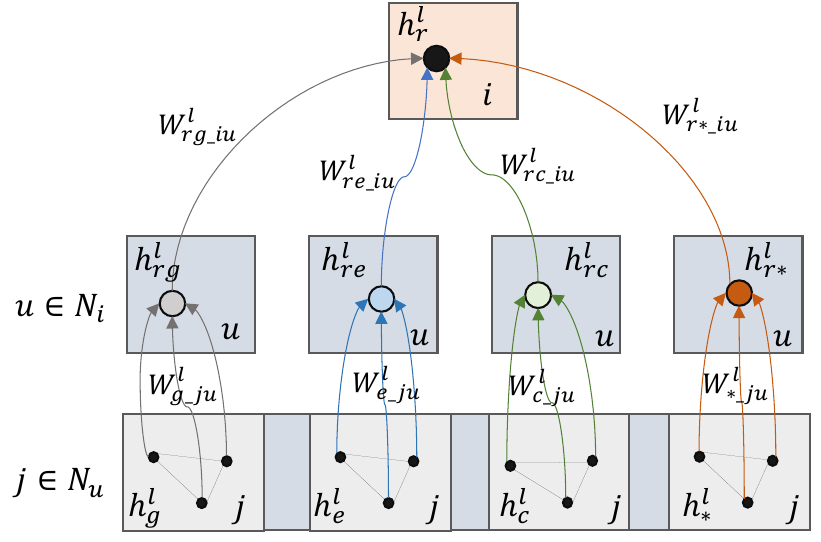}
        \caption{Illustration of node feature update operation in the heterogeneous graph. The node features of the robots at layer $l$ are enhanced through interactions with the other entities in the cross-graphs.}
        \label{fig:operation}
        \vspace{-0.3cm}
    \end{figure}

 \subsubsection{Similarity Measurement} 
    
    We formulate the long-term goal selection problem in multi-robot area search as the graph matching between the robot and candidate graphs. Mathematically, this corresponds to finding a mapping $M: V_r \to V_c$ that maximizes the similarity between the two self-graphs, where $V_r$ and $V_c$ denote the sets of robot nodes and candidate goals, respectively. The mapping function between two self-graphs is constructed by the affinity matrix, and the diagonal and off-diagonal elements of the affinity matrix represent the correlations between the nodes or edges of two self-graphs. The most popular learning-based graph matching is the PCA-GM algorithm proposed in \cite{wang2019learning}. Specifically, the PCA-GM reformulates the graph matching as the linear assignment problem and further operates by the Sinkhorn algorithm. The network is trained using a permutation cross-entropy loss. The Sinkhorn algorithm is used to iterate the affinity matrix to the final corresponding matrix $\boldsymbol{S}$,

    \begin{equation}
        \boldsymbol{S} = Sinkhorn(\boldsymbol{M}^{(0)})
    \end{equation}
    where the $\boldsymbol{M}^{(0)}$ denotes the element of the affinity matrix, and it is normalized on the row and column until convergence. In the final correspondence matrix $\boldsymbol{S}$, the sum of values in each row or column equals 1. Each element $\boldsymbol{S}_{ij}$ represents the matching degree between robot node $V_r$ and candidate node $V_c$. The matching can be completed by performing the argmax operation on each row and column. The correspondence matrix is used to assign the long-term goal from the candidate goals to each robot.
    
\subsection{Action Planner and Execution}

    After obtaining the long-term goal from the topological planner module, the action planner guides each robot to move toward the long-term goal by the short-term path planning algorithm proposed in \cite{ye2022multi}. This path planning algorithm calculates the shortest path from the robot's current position to the long-term goal using the Fast Marching Method (FMM) and considers a safe moving trajectory that avoids collisions with obstacles during the planning process.

    The shortest path from the current position of the robot toward the long-term goal is calculated based on the global occupancy map, and the path is then converted into primitive actions for execution by the robot system. During the conversion process, the shortest path is firstly sub-sampled according to a certain density and generates more discrete points, then selects the nearest points to the robot as the short-term goal. The robot will choose the primitive action from the action space $A = \{Move\mbox{-}forward, Turn\mbox{-}right, Turn\mbox{-}left\}$, and control its rotation and translation by the heuristic algorithm \cite{chen2021topological}. Specifically, if the robot is facing the short-term goal, it will move forward; otherwise, it will perform a designated rotation action.


\subsection{Policy Optimization}

    We train the policy of the multi-robot area search task by the PPO algorithm with the actor-critic architecture, the policy network is parameterized by $\pi_{\theta}$. The actor network is used to train the centralized policy for selecting the long-term goal from the candidate goals. The critic network calculates the state value function and evaluates the rationality of the output from the actor network. The policy gradient ascent follows the equation:

    \begin{equation}   
        \hat{g} =\hat{\mathbb{E}}_t[ \nabla_{\theta} \log \pi_{\theta}(a_{t} \mid s_{t})\hat{A}] 
    \end{equation}

    where each robot shares the centralized policy $\pi_{\theta}$, the $\hat{A} = \sum_{t=1}^{\infty} (\gamma\lambda)^l \delta(t+l)$ denotes the GAE function of the policy, the TD-error is $\delta(t) = r_t + \lambda V_{\phi}(s_{t+1}) - V_{\phi}(s_{t})$, the state value $V_{\phi}(s_{t})$ is predicted by the critic network parameterized by $\phi$, the $r_t$ denotes the global reward feedback received from the environment after executing action $a_t$. For the training policy, we define three distinct rewards for all robots: exploration reward $R^e$, coverage reward $R^c$, and the time penalty $\alpha_3$. The exploration reward $R_t^e$ at timestep $t$ is the difference in the explored area $S^e$ between timestep $t$ and timestep $t-1$ for all robots. The coverage reward is the total number of located targets at timestep $t$, which denotes $S^c_t$.

    \begin{equation}
        r_t = \alpha_1(S^e_t-S^e_{t-1}) + \alpha_2 S^c_t + \alpha_3
    \end{equation}
    
    By setting three distinct rewards, the robots can be guided to quickly explore unknown areas, discover and locate more targets, and avoid wasting excessive time and resources. This realizes the balance among multiple tasks and objectives during the task and optimizes overall task execution efficiency. Building on this, the objective of the proposed policy in terms of reinforcement learning is represented as follows:

    \begin{equation}
        L^{PPO} = L^{CLIP} + c_1 L^{KL} + c_2 L^{VF} + c_3 L^{E}
    \end{equation}

    where $L^{CLIP}$, $L^{KL}$, $L^{VF}$, and $L^{E}$ are the clipped surrogate function, KL penalty, value function error of the policy, and entropy bones respectively; $c_1$, $c_2$, and $c_3$ are coefficients of KL penalty, value function and entropy.

    In addition to pursuing the maximization of the accumulated global reward, we also enhance the PPO algorithm with a prediction loss function that improves the accuracy of perception information representation. Concretely, the prediction network receives the state information $s_t$ at timestep $t$, then predicts the exploration ratio ${y}_t$. The mean squared error between the predicted exploration ratio and ground truth is calculated to train the prediction network:

    \begin{equation}
        L^{Acc} = \frac{1}{T} \sum_{k=1}^T (y_t-\hat{y}_t)^2
    \end{equation}
    
    The final loss function during training can be represented as a weighted sum of the exploration rate prediction loss function $L^{Acc}$ and the PPO loss function $L^{PPO}$.

    \begin{equation}
        L^{PPO} = L^{CLIP} + c_1 L^{KL} + c_2 L^{VF} + c_3 L^{E} + c_4 L^{Acc}
    \end{equation}
    where the weight parameters $c_4$ that decrease with the training step increase, ensuring that the exploration rate prediction loss function mainly assists in the early stages of training. As training progresses, its influence gradually diminishes, allowing network training to be primarily driven by reinforcement learning in the later stages.


    



\section{Experiments}

    In this section, we conduct experiments to answer the following questions: (1) How does the efficiency of our method compare to existing research? (2) What about the generalization of our method? Can our method be extended to more complex environments? (3) How does the scalability of our method? Can our method be applied to scenarios involving a larger number of robots? The well-designed experiments validate the effectiveness of our method.

\subsection{Experiment Setup}
\begin{figure}[!t]
        \vspace{0.5em}
        \centering
        \includegraphics[scale=0.5]{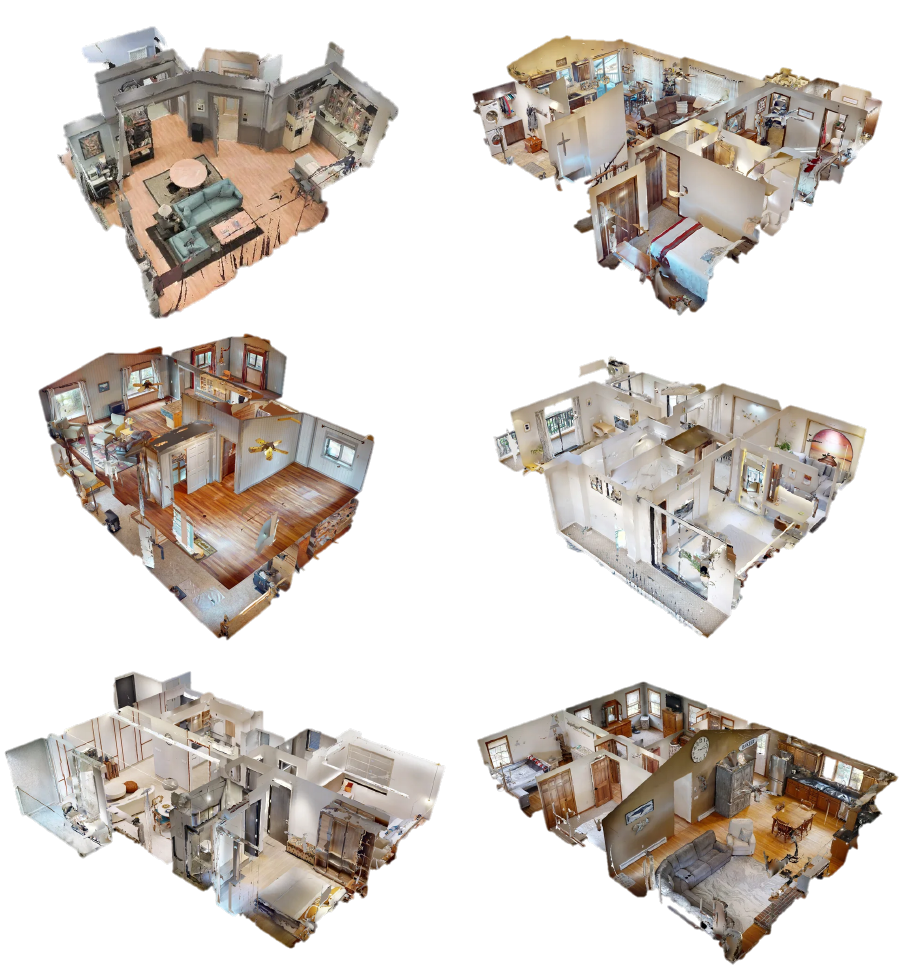}
        \caption{The scenes of the Gibson and Matterport3D datasets, which are collected from real indoor spaces. The robots can obtain egocentric RGBD images and their corresponding positions in a simulation scene.}
        \label{fig:simulation}
        \vspace{-0.3cm}
    \end{figure}

    We utilize the iGibson simulator with the Turtle robot model to generate the experimental scenes. The validated dataset of our proposed method consists of the Gibson and MatterPort3D dataset, as shown in \ref{fig:simulation}. We filter out some scenes where the area sum is too low or the robots couldn't attain the $10\%$ area of the entire scene. We follow \cite{ye2022multi} by selecting 9 scenes from the Gibson dataset for training, sampling the other 44 scenes from the Gibson dataset and 48 scenes from the MatterPort3D dataset for testing. To better analyze the performance in terms of the scene scale, we split the chosen scenes into three parts: small scenes, middle scenes, and large scenes. The small scenes with whole areas ranging from $0m^2$ to $31m^2$, the middle scenes with whole areas ranging from $36m^2$ to $66m^2$, and the large scenes with whole areas ranging from $69m^2$ to $1000m^2$. We randomly generate 100 targets in the free spaces of scenes with various scales. We perform RL training with $12\times10^6$ timesteps. The testing result of each scene is an average of over 50 testing episodes.

    \textbf{Network Architecture and Hyperparameter Details:} Our proposed method is trained using an actor-critic architecture with the PPO algorithm. The critic network is composed of an embedding network and a state value function network. The embedding network with a 5-layer CNN architecture extracts the 32-dimensional feature vector from the observation map. This feature vector is then fed into the state value function network, which consists of 3 fully connected layers and outputs the 1-dimensional state value. The actor network is built upon a graph embedding and aggregation network. The graph embedding network encodes the local information into the node feature vectors for the graph neural network. It comprises a common encoder, a special encoder, and a distance encoder. The common encoder with 5 fully connected layers extracts the 64-dimensional feature vector from the frontier and target cells. The distance encoder with a single fully connected layer captures the high-dimensional spatial relation information between the robots and candidate goals from the distance map. The special encoder with double 5 fully connected layers extracts the 32-dimensional structural and trajectory features from the historical map. The extracted embedding feature vectors serve as the initial node features of the graph. These features are subsequently aggregated using a graph attention network, which employs six iterations of multi-head attention or MLP attention on the self-graph and cross-graph, respectively.
    
    Our model was trained using an Intel(R) Core(TM) i9-12900K CPU and an Nvidia GeForce RTX 4080 GPU in Pytorch, taking approximately 62 hours to complete $5\times12^6$ timesteps. The 9 parallel threads are employed through the Pytorch, the single scene occupies one parallel. In all training algorithm configurations, the optimization is conducted with the learning rate of $5\times 10^{-4}$, the PPO clipping parameter $\varepsilon\mbox{-}greedy$ is set to 0.2, and the discount factor is set to 0.99. 
    The topological mapper samples a new long-term goal every 40 local timesteps for each global timestep, and sets 120 global timesteps within each episode.

    \textbf{Evaluation Metrics:} We evaluate the performance of our method in comparison with the state-of-the-art methods in terms of percentage and efficiency of task completion. The percentage of task completion is related to the sub-tasks of multi-robot area search, encompassing two parts: exploration and coverage percentage. The efficiency in executing these sub-tasks demonstrates the collaboration among robots. Below are the specific evaluation metrics:

    1) \emph{Exploration percentage (Explo ($\%$))}: At each timestep, the explored area of the environment equals the union of the sensor ranges of all robots. The total explored area per episode sums the explored area from the initial test timestep to the maximum test timestep. The exploration percentage is calculated by averaging the explored area over 500 test episodes, divided by the entire free area excluding obstacle cells.

    2) \emph{Coverage percentage (Cover ($\%$))}: This metric represents the cumulative covered targets divided by all coverable targets. Only the area that has been explored allows the robot to locate the targets randomly generated within it. Thus, this metric heavily relies on the exploration percentage.
    
    3) \emph{Time efficiency}: This metric measures the average consumed timesteps required to explore and cover 80\% of the entire unknown environment, denoted as $Time$\_$E (step)$ and $Time$\_$C (step)$, respectively.



\subsection{Baseline Methods}

    We verified the superiority of our method through a comprehensive comparison with various baseline methods in the well-designed experiments. The baseline methods encompass three distinct types: heuristic methods, expert methods, and state-of-the-art (SOTA) methods. Specifically: 

     We extend this baseline algorithm applied to the exploration task to the coverage and exploration task by adding the target cells to the candidate frontier sets.

    \textbf{Utility:} It is a utility-maximizing method \cite{julia2012comparison}. Each robot selects the most frontier cells as its long-term goal, weighing the utility and exploration cost functions of all frontier cells.

    \textbf{Greedy:} We extend the exploration problem of the Nearest Frontier approach \cite{visser2013discussion} to the coverage and exploration task, where each robot selects the nearest cell from the frontier or target cells as its long-term goal.

    \textbf{Voronoi:} This graph search-based algorithm proposed in \cite{bhattacharya2014multi} divided the global map into several Voronoi partitions, and each robot chooses the nearest frontier cells of the closest Voronoi partition as its long-term goal.


    \textbf{Coscan:} This method \cite{dong2019multi} 
    dynamic assigns the long-term goals for all robots based on optimal mass transport optimization and plans a smooth path over its assigned frontier goal by solving the TSP problem.

    \textbf{NCM:} \cite{ye2022multi} It formulates the assignment problem between the frontier nodes and robots as neural bipartite graph matching, and predicts the neural distance between the two types of nodes based on the multiple graph neural network.

\subsection{The Efficiency and Generalization Analysis}

\begin{figure*}[htb]
        \centering
         \subfloat[]{\includegraphics[scale=0.5]{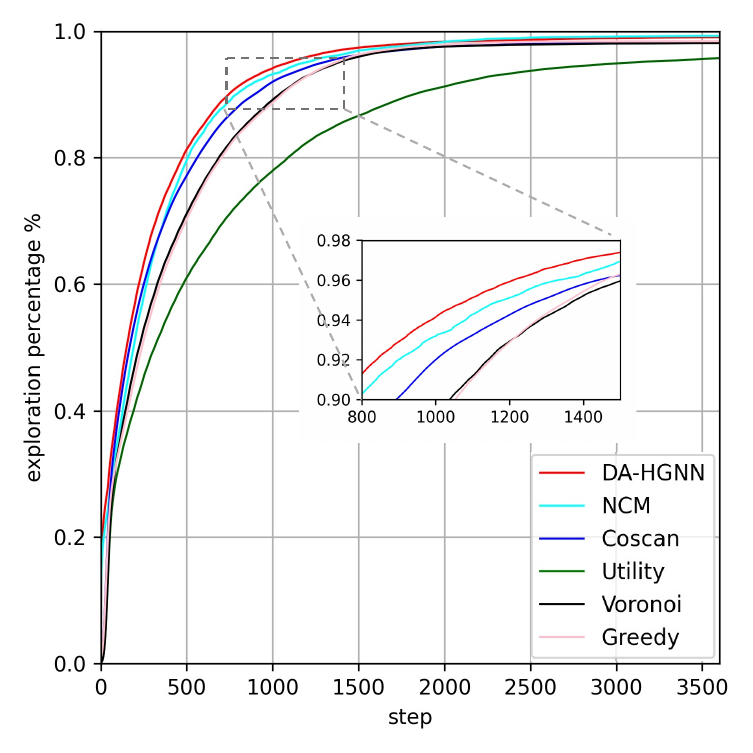}
        \label{fig:test_in_exploration}}
        \hfil
        \subfloat[]{\includegraphics[scale=0.5]{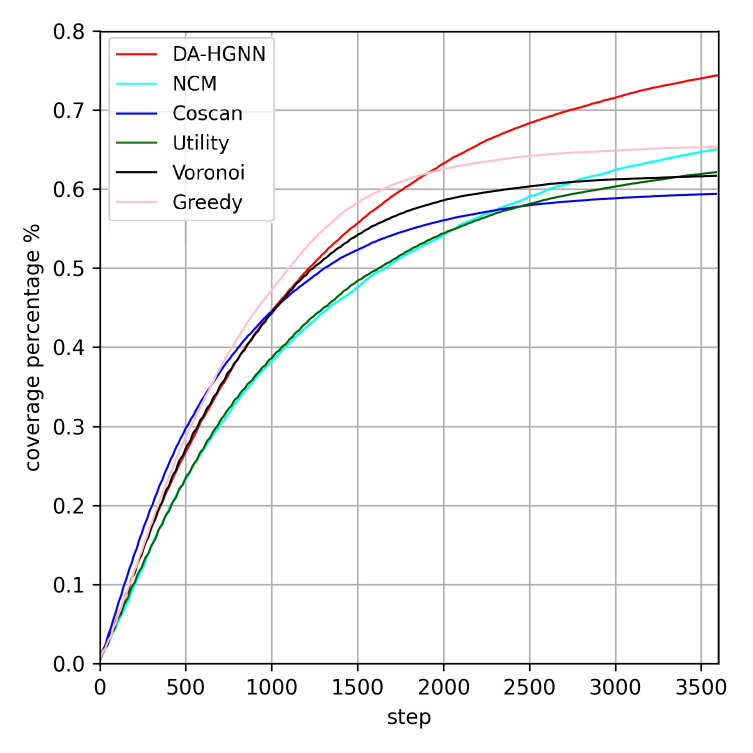}
        \label{fig:test_in_coverage}} 
        \caption{Performance of the testing scenes for exploration and coverage, compared with baseline methods. (a) Exploration percentage. (b) Coverage percentage. Both metrics validate the effectiveness of sub-tasks: the exploration percentage demonstrates the exploration capability (shown in (a)), while the coverage percentage illustrates the ability in coverage (shown in (b)).}
        \label{fig:test_resulting}
    \end{figure*}

    We report the efficiency of the proposed method compared with baseline methods from the qualitative results in Fig.\ref{fig:test_resulting}, where all methods tested on the MatterPort3D dataset. During the testing, we deploy three robots in the middle scenes to minimize the impact of the map distribution on the result. Considering the two sub-tasks, we validate the superior performance over all methods regarding the exploration and coverage percentage. The qualitative results show that all methods can achieve a roughly complete exploration despite the Utility method. The exploration rate of the learning-based approaches (NCM and DA-HGNN) outperforms that of the other planning-based approaches, indicating that the learning-based approaches possess superior long-term planning capabilities in multi-robot complex tasks. Besides, we found that only our method achieves superior performance in terms of coverage percentage. Our proposed method simultaneously achieves the best performance over two sub-tasks, which proves that our method exploits the topological planner module to better trade off between exploration and coverage.

    To validate the efficiency and generalization performance of our proposed method from the quantitative results in Table \ref{table:3robots}, we test all methods in the scenes with three scale areas and three robots. The experimental setting is the same as the training phase, the number of targets is set as 100, and the maximum testing step is set as 3600. Among the effective test scenes on the MatterPort3D dataset, the numbers of the small, medium, and large scenes are 13, 15, and 17, respectively. From the results, we observe that our method achieves the highest exploration percentage, coverage percentage, and time efficiency of coverage in the small scenes, while the time efficiency of exploration is less than the Coscan method. The Coscan focuses on the exploration tasks and ignores the coverage tasks in the whole task execution process, which results in the highest exploration performance and the lowest coverage performance. Our method exhibits the most significant superiority compared to all baseline methods in the medium and large scenes, especially in terms of coverage percentage and efficiency. These quantitative results demonstrate the outstanding generalization of our method to larger scenes. Through the vertical comparisons, as the area of scenes increases, the advantage of our method over the baseline (e.g., Utility) in terms of coverage percentage decreases from $68.31\%$ to $4.66\%$. We consider that such results mainly stem from the following aspects. 1) As the size of scenes increases, the candidate goals are relatively scattered within the explored area. Meanwhile, the robot locating its long-term goal from a set of dense candidate goals may consume more timesteps than when selecting from the sparse candidate goals of the small scenes. 2) As the scene size grows, the increased computational cost limits the ability of our method to maintain efficiency, allowing simpler methods to perform better.
    


    \begin{table}[]
        \renewcommand{\arraystretch}{1.2}
        \centering
        \caption{Quantitative Results of Deploying 3 Robots in Different Scenes}
        \label{table:3robots}
        \begin{tabular}{c|c c c c}
            \toprule
            & \multicolumn{4}{c}{Small Scene }\\
            \hline
            Method & Explo(\%) & Time$\_$E (step) & Cover(\%)  & Time$\_$C (step) \\
            \hline
            Utility & 98.82 & 646.79 & 45.39 & 2981.57\\
            Greedy & 98.90 & 452.47 & 45.51 & 2922.96\\
            Voronoi & 98.48 & 417.39 & 42.82 & 3047.90 \\
            Coscan  & 98.43 & \textbf{382.62} & 39.51 & 3167.36   \\
            NCM & 98.67 & 530.39 & 76.40 & 1768.73\\
            DA-HGNN & \textbf{98.96} & 477.00 & \textbf{77.46} & \textbf{1742.53} \\
        \end{tabular}
       \begin{tabular}{c|c c c c}
            \hline
            & \multicolumn{4}{c}{Medium Scene }\\
            \hline
          Method & Explo(\%) & Time$\_$E (step) & Cover(\%)  & Time$\_$C (step) \\
            \hline
            Utility & 99.22 & 1207.56 & 62.28 & 2525.83\\
            Greedy & 99.14 & 860.04 & 63.84 & 2281.85\\
            Voronoi & 99.14 & 863.51 & 61.02 & 2429.04 \\
            Coscan  & 99.00 & 776.64 & 58.04 & 2507.61  \\
            NCM & 99.30 & 734.30 & 64.38 & 2565.04\\
            DA-HGNN & \textbf{99.61} & \textbf{650.07} & \textbf{75.75} & \textbf{2077.03} \\
        \end{tabular}
       \begin{tabular}{c|c c c c}
            \hline
            & \multicolumn{4}{c}{Large Scene }\\
            \hline
            Method & Explo(\%) & Time$\_$E (step) & Cover(\%)  & Time$\_$C (step) \\
            \hline
            Utility & 97.86 & 1491.35 & 68.20 & 2475.36\\
            Greedy & 98.49 & 1210.48 & 68.97 & 2213.87\\
            Voronoi & 98.64 & 1199.90 & 66.79 & 2336.14 \\
            Coscan  & 98.17 & 1148.66 & 63.52 & 2455.79  \\
            NCM & 98.77 & 1138.69 & 70.47 & 2235.78\\
            DA-HGNN & \textbf{98.95} & \textbf{1020.00} & \textbf{71.38} & \textbf{2160.04} \\
            \bottomrule
    \end{tabular}
    \end{table}
    
    \begin{table}[]
        \renewcommand{\arraystretch}{1.2}
        \centering
        \caption{Quantitative Results of Deploying 4 Robots in Different Environments}
        \label{table:4robots}
        \begin{tabular}{c|c c c c}
            \toprule
            & \multicolumn{4}{c}{Small Scene }\\
            \hline
            Method & Explo(\%) & Time$\_$E (step) & Cover(\%)  & Time$\_$C (step) \\
            \hline
            Utility & 98.68 & 450.96 & 46.40 & 2868.79\\
            Greedy & 98.60 & 388.56 & 48.14 & 2758.31\\
            Voronoi & 98.62 & 351.06 & 38.86 & 3198.72 \\
            Coscan  & 97.82 & \textbf{268.44} & 37.64 & 3199.90   \\
            NCM & 98.75 & 418.51 & 70.23 & 1918.04 \\
            DA-HGNN & \textbf{99.90} &  539.11 & \textbf{74.72} & \textbf{1852.85} \\
        \end{tabular}
       \begin{tabular}{c|c c c c}
            \hline
            & \multicolumn{4}{c}{Medium Scene }\\
            \hline
          Method & Explo(\%) & Time$\_$E (step) & Cover(\%)  & Time$\_$C (step) \\
            \hline
            Utility & 98.36 & 1075.97 & 59.62 & 2539.76\\
            Greedy & \textbf{99.25} & 770.78 & 67.35 & 2071.33\\
            Voronoi & 99.08 & 656.60 & 51.23 & 2739.45 \\
            Coscan  & 98.87 & \textbf{554.43} & 49.47 & 2777.62  \\
            NCM & 98.88 & 661.86 & 66.55 & 2271.54 \\
            DA-HGNN & 99.64 & 583.47 & \textbf{77.44} & \textbf{1894.84} \\
        \end{tabular}
       \begin{tabular}{c|c c c c}
            \hline
            & \multicolumn{4}{c}{Large Scene }\\
            \hline
            Method & Explo(\%) & Time$\_$E (step) & Cover(\%)  & Time$\_$C (step) \\
            \hline
            Utility & 98.39 & 1384.70 & 71.12 & 2264.30\\
            Greedy & 98.58 & 1108.25 & 67.75 & 2182.93\\
            Voronoi & 98.47 & 972.43 & 54.42 & 2718.41  \\
            Coscan  & 98.07 & 871.76 & 50.93 & 2835.57  \\
            NCM & 97.99 & 1005.58 & 57.59 & 2792.11\\
            DA-HGNN & 98.88 & \textbf{979.44} & \textbf{74.17} & \textbf{2001.73} \\
            \bottomrule
    \end{tabular}
    \end{table}

    \begin{table}[]
        \renewcommand{\arraystretch}{1.2}
        \centering
        \caption{Quantitative Results of Deploying 5 Robots in Different Environments}
        \label{table:5robots}
        \begin{tabular}{c|c c c c}
            \toprule
            & \multicolumn{4}{c}{Small Scene }\\
            \hline
            Method & Explo(\%) & Time$\_$E (step) & Cover(\%)  & Time$\_$C (step) \\
            \hline
            Utility & 98.43 & 431.35 & 44.34 & 2951.89 \\
            Greedy & 98.81 & 374.08 & 47.98 & 2765.36 \\
            Voronoi & 98.34 & 319.08 & 40.03 & 3111.98 \\
            Coscan & 98.31 & \textbf{254.82} & 37.61 & 3206.39 \\
            NCM & \textbf{99.17} & 326.03 & 74.74 & 1662.25 \\
            DA-HGNN & 98.54 & 392.54 & \textbf{81.25} & \textbf{1225.33} \\
        \end{tabular}
       \begin{tabular}{c|c c c c}
            \hline
            & \multicolumn{4}{c}{Medium Scene }\\
            \hline
          Method & Explo(\%) & Time$\_$E (step) & Cover(\%)  & Time$\_$C (step) \\
            \hline
            Utility & 99.12 & 938.02 & 63.24 & 2320.05 \\
            Greedy & 99.39 & 711.46 & 67.71 & 2008.23 \\
            Voronoi & 98.97 & 591.67 & 54.04 & 2565.55 \\
            Coscan & 99.02 & \textbf{502.26} & 50.08 & 2723.07 \\
            NCM & 99.32 & 578.21 & 73.24 & 1949.65 \\
            DA-HGNN & \textbf{99.66} & 541.64 & \textbf{79.56} & \textbf{1502.98} \\
        \end{tabular}
       \begin{tabular}{c|c c c c}
            \hline
            & \multicolumn{4}{c}{Large Scene }\\
            \hline
            Method & Explo(\%) & Time$\_$E (step) & Cover(\%)  & Time$\_$C (step) \\
            \hline
            Utility & 98.01 & 1308.43 & 68.73 & 2267.77 \\
            Greedy & 98.72 & 997.61 & 70.18 & 2006.36 \\
            Voronoi & 98.25 & 840.43 & 56.02 & 2589.27 \\
            Coscan & 98.35 & \textbf{803.63} & 52.54 & 2736.98 \\
            NCM & 98.25 & 885.76 & 59.47 & 2659.51 \\
            DA-HGNN & \textbf{98.90} & 848.42 & \textbf{77.87} & \textbf{1739.60} \\
            \bottomrule
    \end{tabular}
    \end{table}

\subsection{The Efficiency and Scalability Analysis}

   We further verify the efficiency and scalability of our proposed method across various scene scales, where the number of robots in the testing scenes is extended from 3 to 4 and 5, respectively. The experimental results are shown in Table \ref{table:3robots}, \ref{table:4robots}, and \ref{table:5robots}. 
   
   As we can see, the performance of our proposed adaptive decision-making algorithm outperforms all baseline methods across testing scenes with varying numbers of robots in terms of exploration percentage, coverage percentage, and coverage efficiency. In the small scenes, although the Coscan takes a marginal advantage in the consumed time step of exploration, the DA-HGNN shows more prominent overall performance, achieving the highest exploration percentage of $99\%$. This implies that the proposed method can more comprehensively ensure efficient exploration within the restricted scenes. Besides, the DA-HGNN excels in the consumed time step of coverage among all compared methods, which means it achieves broader and more efficient target coverage within the limited timestep, thereby accelerating task completion. In both medium and large scenes, whether deploying 4 or 5 robots, the DA-HGNN consistently achieves the highest exploration and coverage percentage. The stable performance demonstrates the ability to adapt to varying scene scales flexibly. Through rational target allocation and path planning, it effectively reduces the overall coverage time, enabling rapid and efficient scene exploration and area coverage.

   The reason DA-HGNN maintains scalability across all scales of testing scenes stems from its topological planning with type-aware graph augmentation, which ensures a well-balanced coordination between the two sub-tasks at each timestep during the decision-making process. The relational-aware and type-aware graph augmentation sub-modules analyzes the topological and semantic relationships between the robots and the candidate target nodes over the heterogeneous graph. This process incorporates not only the spatial topological relations between two nodes, but also the semantic information regarding two nodes. The topological and semantic information is key to the DA-HGNN maintaining its effectiveness and scalability across scenes with deploying a varying number of robots.

        \begin{table}[]
        \renewcommand{\arraystretch}{1.2}
        \centering
        \caption{Ablation Study on the Node Semantic Feature and History Feature}
        \tabcolsep=0.1cm  
        \label{table:ablation_study}
        \begin{tabular}{c|c c c c}
            \toprule
            & \multicolumn{4}{c}{Small Scene }\\
            \hline
            Method & Explo(\%) & Time$\_$E (step) & Cover(\%)  & Time$\_$C (step) \\
            \hline
            w/o type semantic & 99.00 & 499.50 & 77.12 & 1742.82\\
            w/o history module & 99.04 & 488.24 & 77.23 & 1744.99 \\
            DA-HGNN & \textbf{98.96} & \textbf{441.92} & \textbf{80.29} & \textbf{1563.86} \\
        \end{tabular}
       \begin{tabular}{c|c c c c}
            \hline
            & \multicolumn{4}{c}{Medium Scene }\\
            \hline
          Method & Explo(\%) & Time$\_$E (step) & Cover(\%)  & Time$\_$C (step) \\
            \hline
             w/o type semantic & 99.56 & 641.99 & 74.22 & 2130.28\\
            w/o history module & 99.60 & 646.33 & 74.33 & 2130.49 \\
            DA-HGNN & \textbf{99.75} & \textbf{603.52} & \textbf{77.53} & \textbf{2029.01} \\
        \end{tabular}
       \begin{tabular}{c|c c c c}
            \hline
            & \multicolumn{4}{c}{Large Scene }\\
            \hline
            Method & Explo(\%) & Time$\_$E (step) & Cover(\%)  & Time$\_$C (step) \\
            \hline
            w/o type semantic & 98.91 & 1071.31 & 71.86 & 2513.09 \\
            w/o history module & 98.91 & 1063.98 & 71.21 & 2528.28  \\
            DA-HGNN & \textbf{98.97} & \textbf{1014.00} & \textbf{72.98} & \textbf{2406.35} \\
            \bottomrule
    \end{tabular}
    \end{table}

\subsection{The Ablation Analysis on Efficacy}

    We conduct an ablation study to investigate the importance of each component of our proposed algorithm to the multi-robot area search problem. The experimental results on three scales of scenes from MatterPort3D datasets are shown in Table \ref{table:ablation_study}. Specifically, we justify the design of nodes type semantic features and the history module by removing them separately from the entire framework for an ablation study. We denote two variants as \textbf{w/o type semantic} and \textbf{w/o history module}, respectively. As we can see that the performance of both variants is worse in percentage and task completion of the two sub-tasks. More details, without the enhancement of the node type feature, the similarity measurement of topological planning is only based on the feature embedding constructed by the topological information between the robots and candidate targets. The process of assigning targets to each robot considers the relative distance between robots and targets, not the semantics of targets. Besides, with the increase of scene size, the search efficiency of the \textbf{w/o type semantic} gradually decreases. This is attributed to the growing importance of the distance feature over node type features, enabling the robots to save substantial timesteps and concentrate more efficiently on exploration and coverage. Then, a comparison of results between the \textbf{w/o history module} and the proposed method reveals the performance decline across all scales of scenes. This demonstrates that historical information enriches local perception information, thereby enhancing environmental representation ability, ultimately improving the decision-making capability of multi-robot systems.

     \begin{table}[]
        \renewcommand{\arraystretch}{1.2}
        \centering
        \caption{The Experimental Analysis on the Utility Method with Different Probability Settings}
        \label{table:probability_baseline}
        \begin{tabular}{c|c c c c}
            \toprule
            & \multicolumn{4}{c}{Small Scene }\\
            \hline
            Method & Explo(\%) & Time$\_$E (step) & Cover(\%)  & Time$\_$C (step) \\
            \hline
            Utility\_{40} & 98.36 & 561.24 & 45.57 & 2932.27\\
            Utility\_{50} & 98.72 & 500.07 & 44.11 & 2991.23\\
            Utility\_{60} & 98.82 & 646.79 & 45.39 & 2981.57 \\
            DA-HGNN & \textbf{98.96} & \textbf{477.00} & \textbf{77.46} & \textbf{1742.53} \\
        \end{tabular}
       \begin{tabular}{c|c c c c}
            \hline
            & \multicolumn{4}{c}{Medium Scene }\\
            \hline
          Method & Explo(\%) & Time$\_$E (step) & Cover(\%)  & Time$\_$C (step) \\
            \hline
            Utility\_{40} & 99.23 & 1093.23 & 63.99 & 2425.74 \\
            Utility\_{50} & 99.07 & 1150.42 & 59.79 & 2600.78\\
            Utility\_{60} & 99.22 & 1207.56 & 62.28 & 2525.83 \\
            DA-HGNN & \textbf{99.61} & \textbf{650.07} & \textbf{75.75} & \textbf{2077.03} \\
        \end{tabular}
       \begin{tabular}{c|c c c c}
            \hline
            & \multicolumn{4}{c}{Large Scene }\\
            \hline
            Method & Explo(\%) & Time$\_$E (step) & Cover(\%)  & Time$\_$C (step) \\
            \hline
            Utility\_{40} & 98.12 & 1482.26 & 65.79 & 2562.27\\
            Utility\_{50} & 98.14 & 1403.07 & 61.54 & 2691.43\\
            Utility\_{60} & 97.86 & 1491.35 & 68.20 & 2475.36  \\
            DA-HGNN & \textbf{98.95} & \textbf{1020.00} & \textbf{71.38} & \textbf{2160.04} \\
            \bottomrule
    \end{tabular}
    \end{table}

\subsection{Impact Analysis of Probability Setting}

    The planning-based method is proposed to address the exploration task. To fairly compare these methods with our proposed method, we extend these methods to multi-robot area search tasks, where exploration and coverage tasks are executed simultaneously. Specifically, the planning-based algorithm calculates the most beneficial frontier node from the candidate frontier nodes for each robot, while calculating the most beneficial interesting node from a set of candidate interesting nodes. The final node is then selected from two types of nodes with a certain probability, $p_s$, as the long-term goal. We set the $p_s$ as $60\%$ in the validated experiment of generalization and scalability. In this analysis, we evaluate the impact of different probability settings ($40\%, 50\%$, and $60\%$) on the performance of the baseline method. Taking the Utility method as an example, we test the Utility variants and our method on the MatterPort3D dataset in Table \ref{table:probability_baseline}. The results demonstrate that our method consistently outperforms the baseline across all metrics and scene sizes. This indicates that regardless of the chosen probability, the baseline method does not surpass our proposed method. Our method exhibits a strong capability for long-term planning and effectively balancing the trade-off between exploration and coverage tasks.

\subsection{Visualization Analysis}

    \begin{figure*}[!t]
        \vspace{0.5em}
        \centering
        \includegraphics[scale=0.45]{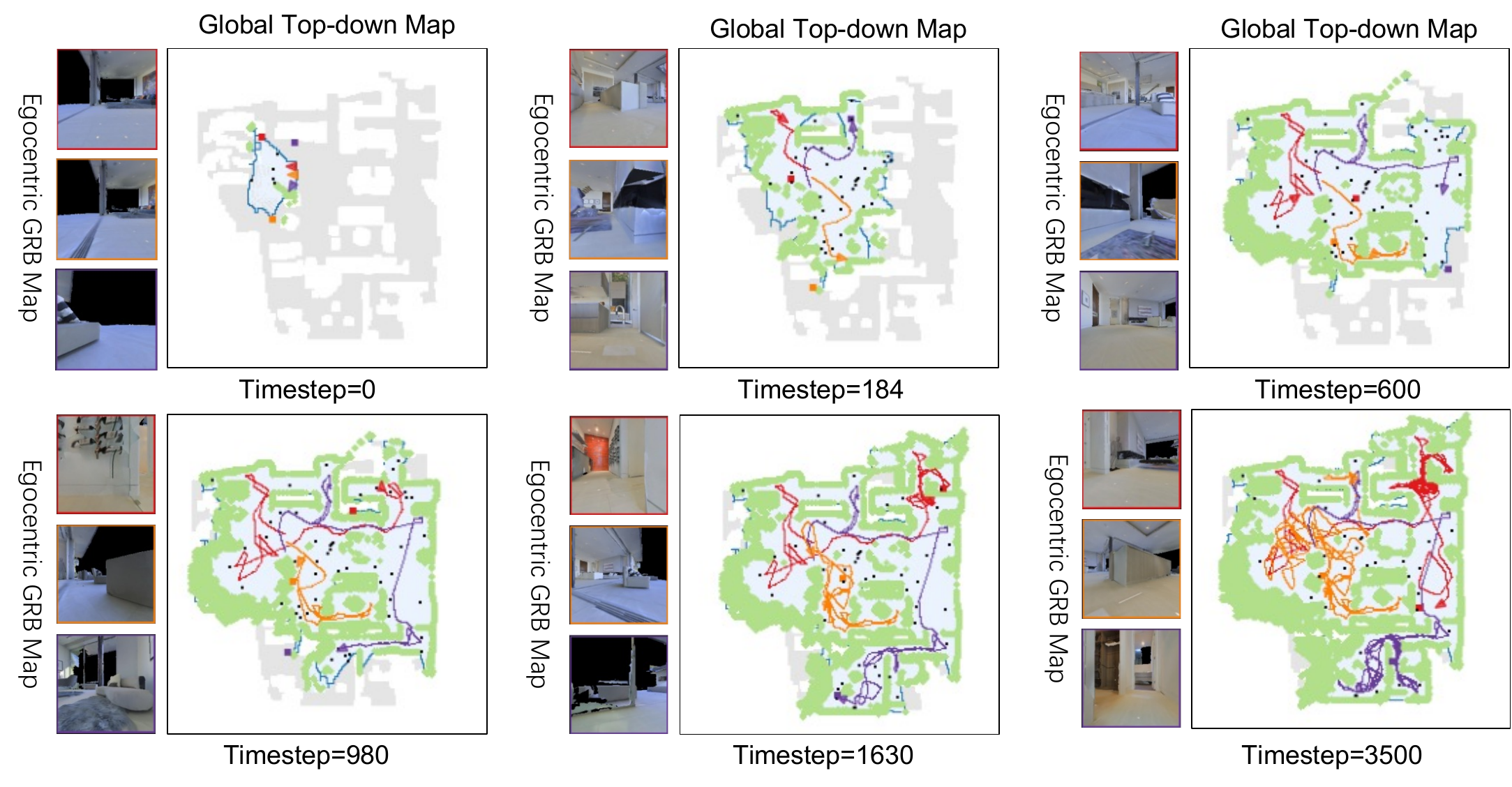}
        \caption{Visualization result in the simulation scene "5LpN3gDmAk7". We deploy three robots in the testing scene, and three colors denote the moving trajectories of the three robots.}
        \label{fig:visualization}
        \vspace{-0.3cm}
    \end{figure*} 

    Fig. \ref{fig:visualization} shows the visualization of our DA-HGNN algorithm's search process in the medium scene samples from the MatterPort3D dataset, denoted by the identifier "5LpN3gDmAk7". The triangle, square, and green circle represent the robots, long-term goals, and targets, respectively. The blue nodes boundary between the unexplored area and the explored area represents the frontier nodes. The black node is the randomly generated target. As we can see, the robots explore and achieve coverage simultaneously in the area search process. This means the long-term goals are selected in a balanced way between the frontier nodes and target nodes. At timestep 1630, when the explored area reaches nearly 100\%, the number of uncovered targets decreases sharply. Moreover, as 70\% of the targets are covered, the limited overlap in robots' trajectories indicates low redundant exploration among the robots. These results stem from the decision-making mechanism, which ensures that when the explored area is sufficiently large, the robots focus more on selecting targets. Our proposed method demonstrates advantages in balancing between frontier nodes and target nodes.


\section{Conclusion and Discussion}

In this work, we develop a Dual-Attention Heterogeneous Graph Neural Network (DA-HGNN) to enhance the dynamic adaptation of the multi-robot area search. We first construct a relational-aware spatio-temporal graph to capture structural dependencies among robots and candidate goals. Then augments the graph with explicit type-aware semantics by differentiating between frontier nodes and target nodes. Through integrated similarity, the method dynamically assigns long-term goals by balancing relational topology and task-specific semantics. Extensive experiments conducted in interactive 3D scenarios within the iGibson simulator, leveraging the Gibson and MatterPort3D datasets, validate the superior scalability and generalization capability of the proposed approach. Comparative results demonstrate statistically significant improvements over state-of-the-art baselines.



\bibliographystyle{IEEEtran}
\bibliography{bare_jrnl_new_sample4.bib}


\vspace{-0.5cm}
\begin{IEEEbiography}[{\includegraphics[width=1in,height=1.25in,clip,keepaspectratio]{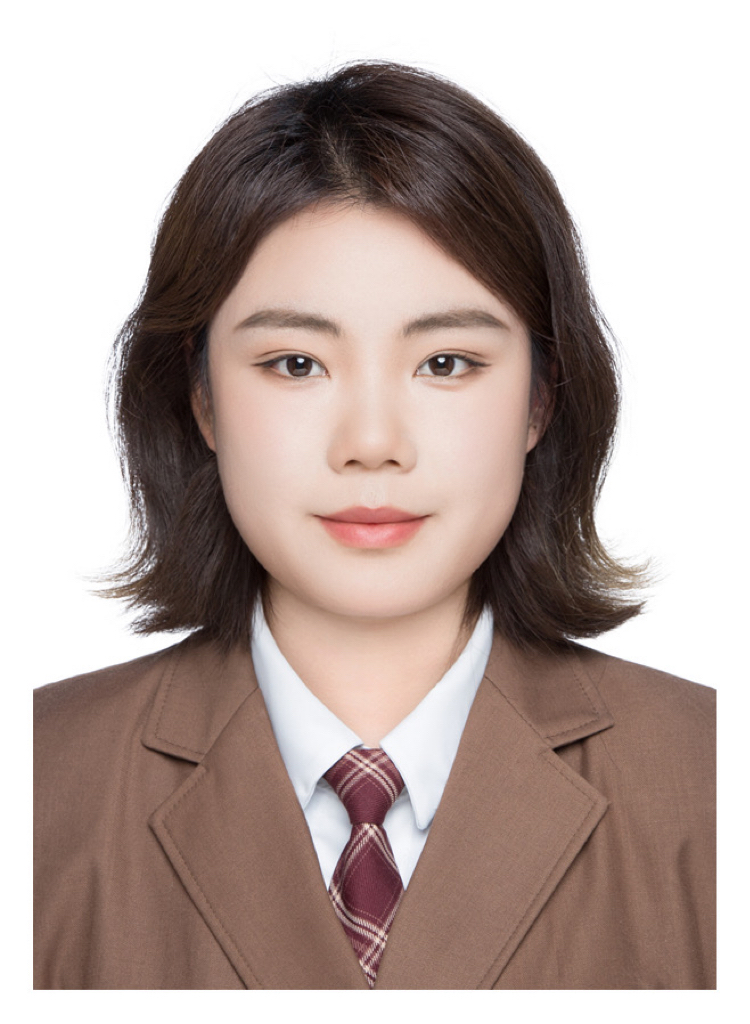}}]{Lina Zhu}
received the Ph.D. degree in control science and engineering from the Institute of Artificial Intelligence and Robotics, Xi'an Jiaotong University, China. 

She is currently with the School of Artificial Intelligence, School of Automation, Xi'an University of Posts \& Telecommunications, Xi'an. Her research interests include multi-robot cooperative sequential decision-making and multi-agent reinforcement learning.
\end{IEEEbiography}

\vspace{-0.5cm}
\begin{IEEEbiography}[{\includegraphics[width=1in,height=1.25in,clip,keepaspectratio]{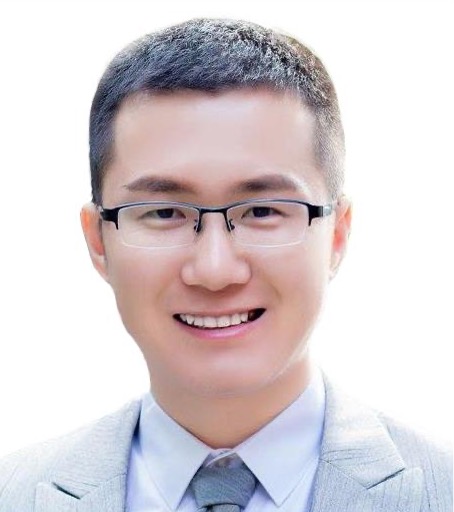}}]{Jiyu Cheng}
received the B.E. degree in automation from Shandong University, Jinan, China, in 2015, and the Ph.D. degree in electronic engineering from The Chinese University of Hong Kong, Hong Kong, in 2019.

He is currently an Associate Professor with the Department of Control Science and Engineering, Shandong University. His current research interests include autonomous navigation and multiagent exploration.
\end{IEEEbiography}

\vspace{-0.5cm}

\begin{IEEEbiography}[{\includegraphics[width=1in,height=1.25in,clip,keepaspectratio]{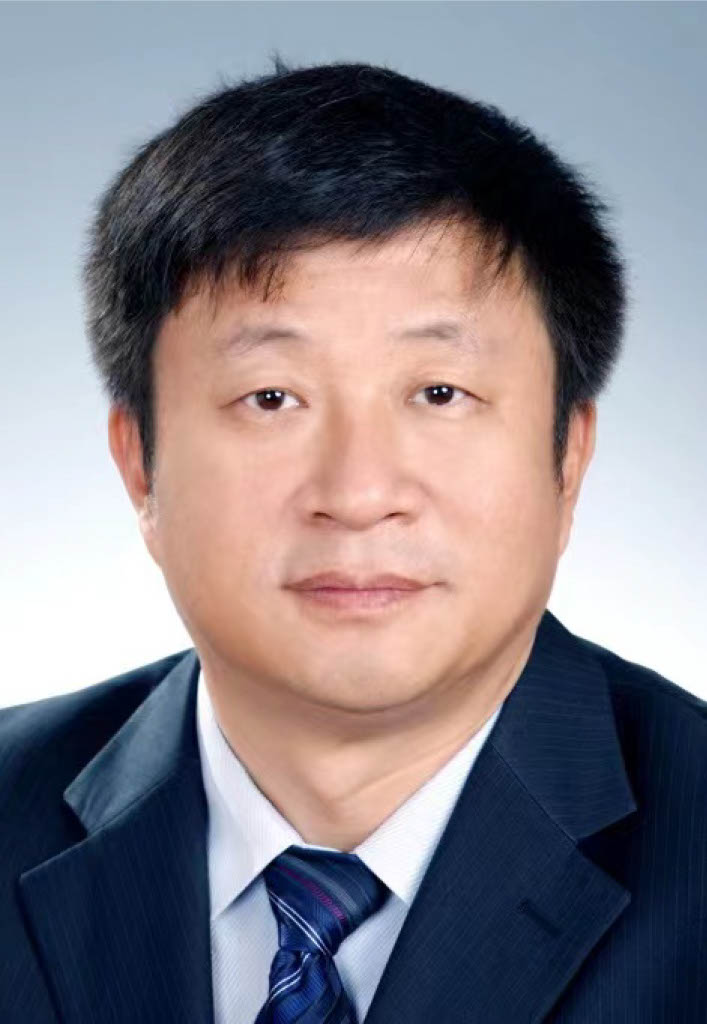}}]{Yuehu Liu}
(Member, IEEE) received the B.E. and M.E. degrees from Xi'an Jiaotong University, Xi'an, China, in 1984 and 1989, respectively, and the Ph.D. degree in electrical engineering from Keio University, Tokyo, Japan, in 2000.

He is currently a Professor with Xi'an Jiaotong University. His current research interests include computer vision, computer graphics, and simulation testing for autonomous vehicle.

Dr. Liu is a member of the IEICE.

\end{IEEEbiography}

\begin{IEEEbiography}
[{\includegraphics[width=1in,height=1.25in,clip,keepaspectratio]{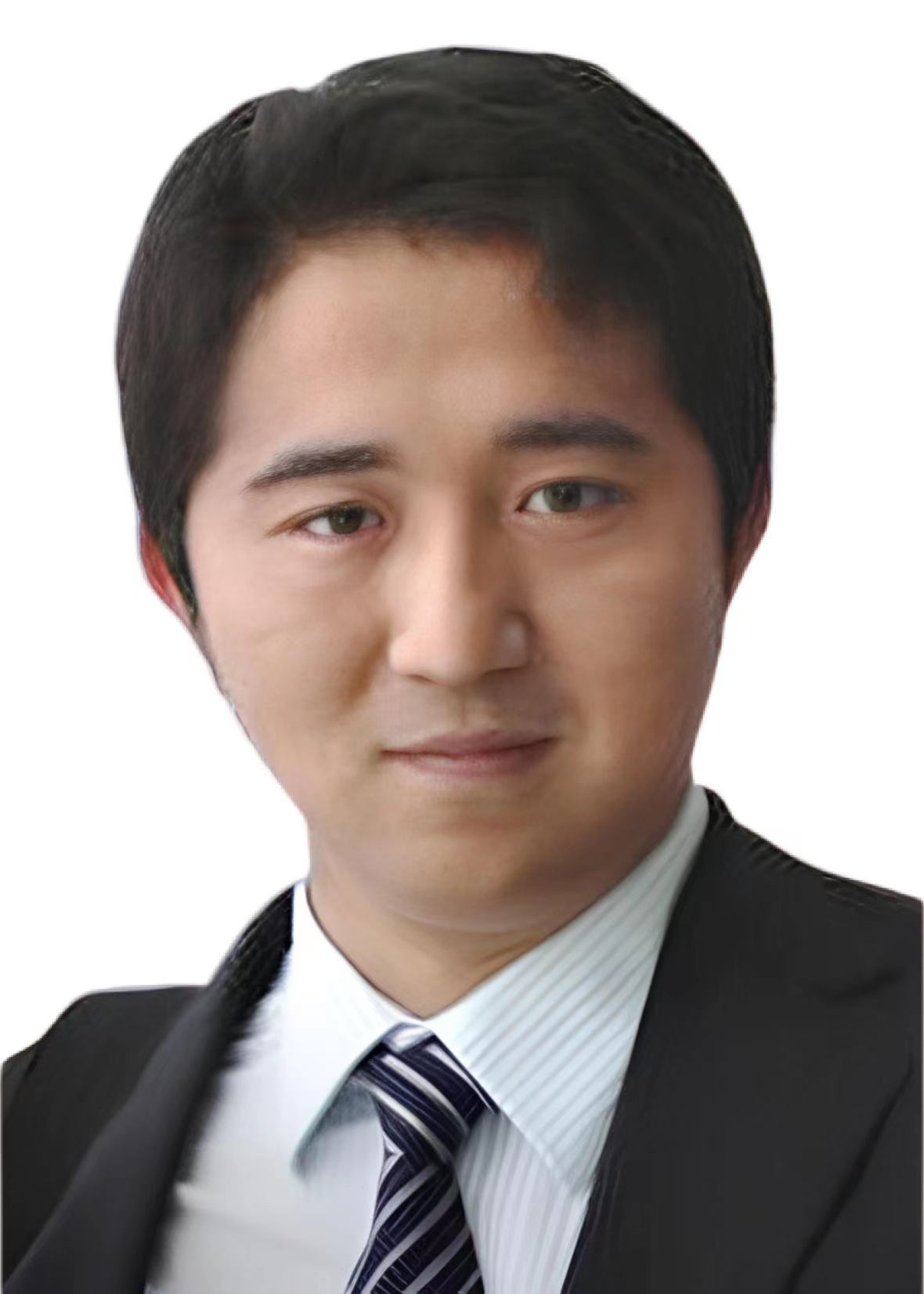}}]{Wei Zhang}
(Member, IEEE) received the Ph.D. degree in electronic engineering from The Chinese University of Hong Kong in 2010.

He is currently with the School of Control Science and Engineering, Shandong University, Jinan, China. His research interests include computer vision and robotics.

Dr. Zhang has served as a program committee member and a reviewer for various international conferences and journals.
\end{IEEEbiography}

\end{document}